\documentclass[runningheads]{llncs}
\usepackage[usenames,dvipsnames]{xcolor} 
\usepackage[utf8]{inputenc}
\usepackage{graphicx}
\usepackage{amsmath}
\usepackage{pdfpages}
\usepackage{booktabs}
\usepackage{color}
\newcommand{\tmpframe}[1]{\fbox{#1}}
\renewcommand{\tmpframe}[1]{#1}

\usepackage[breaklinks=true,bookmarks=false]{hyperref}
\hypersetup{
	colorlinks=true,
	linkcolor=BrickRed,
	citecolor=OliveGreen,
	filecolor=magenta,
	urlcolor=Blue
}
\begin{document}
\title{Segmentation of Head and Neck Organs at Risk Using CNN with Batch Dice Loss}
\titlerunning{Head and Neck Segmentation using CNN with Batch Dice Loss}
%
\author{Oldřich Kodym\inst{1} \and Michal Španěl\inst{1,2} \and Adam Herout\inst{1}}
\authorrunning{Oldřich Kodym et al.}
%
\institute{Graph@FIT, Brno University of Technology,\\
\email{\href{mailto:ikodym@fit.vutbr.cz}{ikodym@fit.vutbr.cz}}
\and
TESCAN 3DIM, Brno, Czech Republic\\ 
\email{\href{mailto:spanel@3dim-laboratory.cz}{spanel@3dim-laboratory.cz}}}

\maketitle              

\begin{abstract}
This paper deals with segmentation of organs at risk (OAR) in head and neck area in CT images which is a crucial step for reliable intensity modulated radiotherapy treatment. We introduce a convolution neural network with encoder-decoder architecture and a new loss function, the batch soft Dice loss function, used to train the network. The resulting model produces segmentations of every OAR in the public MICCAI 2015 Head And Neck Auto-Segmentation Challenge dataset. Despite the heavy class imbalance in the data, we improve accuracy of current state-of-the-art methods by 0.33 mm in terms of average surface distance and by 0.11 in terms of Dice overlap coefficient on average. 

\end{abstract}

\section{Introduction}

Organs at risk (OAR) in head and neck area is a group of organs at potential risk of damage during radiotherapy application. Their three-dimensional segmentation in medical Computed Tomography (CT) images is a first step required for reliable planning in image-guided radiotherapy during head and neck cancer treatment~\cite{Dawson2006}. Producing 3D segmentation in clinical data manually is a tedious task and therefore effort is being put into developing automatic methods that would be able to produce accurate segmentation masks for the objects of interest. In area of head and neck OAR, however, this is a very challenging task as the soft tissue structures have very little contrast.

The MICCAI 2015 Head and Neck Auto Segmentation Challenge~\cite{Raudaschl2017} provides a dataset for evaluation of head and neck OAR segmentation methods. Furthermore, the challenge also defined baseline methods for head and neck segmentation for each structure. Most of the approaches relied on statistical shape model \cite{Heimann2009} or active appearance model \cite{Jung2014} registration with atlas-based initialization during the challenge. Although several methods where able to produce segmentations of every structure, especially smaller objects such as submandibular glands, optic nerves and optic chiasm didn't reach a satisfactory accuracy required for clinical application. More recent methods made use of more modern machine learning methods such as convolutional neural networks (CNN) that have been gaining a lot of popularity since the introduction of AlexNet in 2012~\cite{Krizhevsky2012} in most of computer vision fields including medical image data analysis. 

\begin{figure}[t]
  \centering
  \tmpframe{\includegraphics[width=1\textwidth]{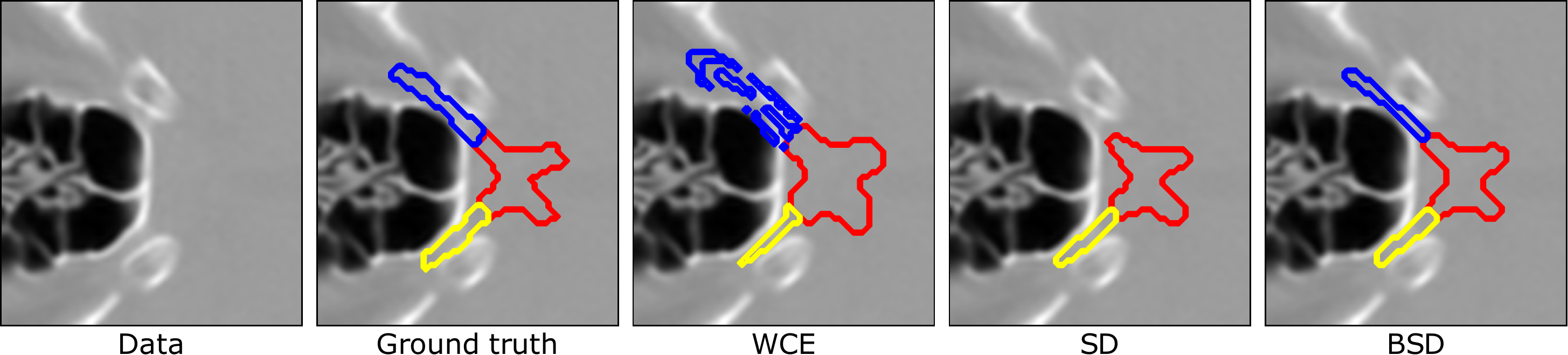}}
  \caption{Examples of optic nerves (blue, yellow) and chiasm (red) segmentation using various standard loss function such as weighted cross-entropy (WCE) and soft Dice (SD) and our Batch soft Dice (BSD) loss.}
  \label{fig:ExamplesOfResults}
\end{figure}

Fritscher et al. \cite{Fritscher2016} first used a patch-based CNN with 3 orthogonal input patches to obtain a pixel-wise prediction map for each structure, using atlas-based probability map for each structure as an additional input, slightly increasing accuracy of parotid and submandibular gland segmentation. Coincidentally, Ibragimov et al. \cite{Ibragimov2017} used a very similar model with a Markov random field post processing step. They obtained segmentation of all OARs but reaching accuracy of only $37\,\%$ for optic chiasm and under $65\,\%$ for optic nerves. Most recently, Wang et al. \cite{Wang2018} applied a hierarchical random forest vertex regression method for some of the OAR, showing further improvement of accuracy of brain stem, mandible, and parotid gland segmentation.

In this paper we design a CNN with encoder-decoder architecture, first used in biomedical segmentation by Ronneberger et al. \cite{Ronneberger2015}, and evaluate its performance on head and neck OAR segmentation task. Since the model fails to learn some of the structures when trained using the standard cross-entropy loss, we make use of Dice loss function introduced first by Pastor-Pellicer et al.\cite{Pastor2013} and later used for medical segmentation by Milletari et al.~\cite{Milletari2016}. This form of \emph{soft Dice loss} has been employed quite extensively in recent literature \cite{Sudre2017,Fidon2017}. Despite obtaining acceptable results on most of the structures, we also observed rather low performance on smaller low-contrast structures when using standard Dice loss. We propose a modification to the standard soft Dice loss function -- the \emph{Batch soft Dice Loss} -- to overcome this problem and show that it enables the model to outperform models trained using other loss functions. In case of optic chiasm and nerves segmentation we reach as much as $59\,\%$ improvement over current state-of-the-art methods in terms of Dice overlap measure.

\section{Proposed Method for Organ Segmentation}

In this section, we describe the architecture of our CNN model and different loss functions that have been evaluated. We also give details on the training phase and data preprocessing.


\subsection{Head and Neck Segmentation Challenge}
The dataset includes CT scans of patients with manual segmentations of 6 anatomical structures which include brain stem (BS), mandible (MA), optic chiasm (OC), bilateral optic nerves (ON), parotid glands (PG), and submandibular glands (SG). 
Total of 48 patient scans of head and neck area are available in the challenge dataset. However, 18 of these scans contain incomplete ground truth annotation for some structures. Since our model is trained using image patches that span across almost complete head area, these scans had to be excluded from our experiments to prevent introducing false background voxels into the ground truth. 


\subsection{Model Architecture}
Our segmentation model architecture is of encoder-decoder type \cite{Ronneberger2015}. Convolutional layers are coupled with max-pooling layers to increase the field of view of deeper features while decreasing their resolution in the first, encoder part. In the second, decoder part, the features are upscaled again using bilinear interpolation. Each upscaling step is accompanied by concatenating the feature maps from the encoder part of the model with matching resolution to improve the gradient flow through the model.

We limit our model to only operate on two-dimensional axial slices for the following two reasons. First, because we use $140 \times 140$ image patches to include enough context, there is an intra-image class imbalance issue. Although we compensate for this during the training as mentioned in the following sections, using three-dimensional image patches results in amplification of this issue because some structures, such as optic chiasm, are highly planar in the axial plane. Second, memory requirements are an issue here as well. The nature of multi-class segmentation requires mini-batches of data used in the training phase to contain a balanced number of image patches containing each of the structures in order to correctly compute the gradient step. This is easier to accomplish when using 2D image patches. Our results show that the two-dimensional approach has only small impact on the $z$-dimension discontinuities and the overall performance.

\begin{figure}[t]
  \centering
  \tmpframe{\includegraphics[width=\textwidth]{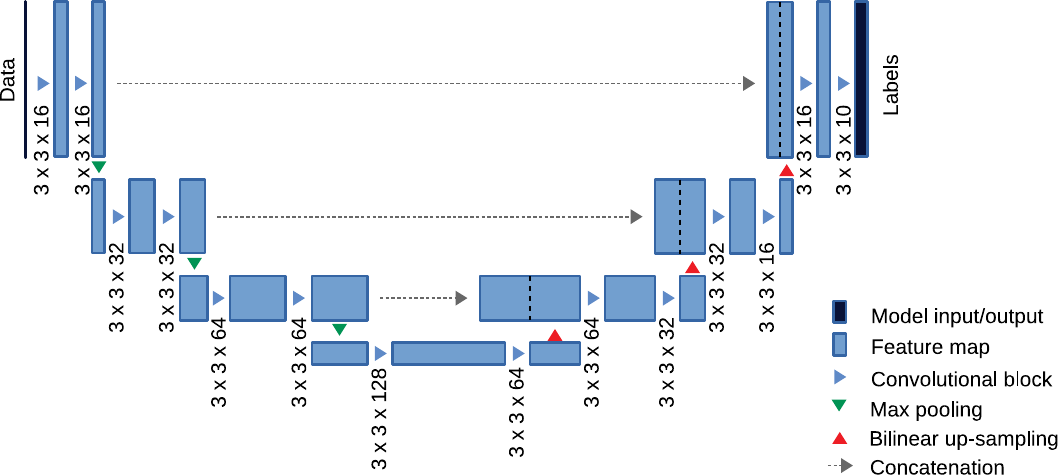}}
  \caption{Proposed segmentation model architecture. Series of two convolutions, batch normalizations and ReLU activations are used in each block. Down- and up-sampling is done using max-pooling and bilinear interpolation, respectively, with concatenation skip connections providing additional spatial information during up-sampling.}
  \label{fig:CNNstructure}
\end{figure}

The architecture scheme is shown in Figure \ref{fig:CNNstructure}. We only use standard $3~\times~3$ convolution kernel size. Each convolutional block encompasses convolution kernel filtering, batch normalization, and ReLU activation, except for the last convolutional block which uses softmax activation to produce the final label probabilities. Unlike U-net, we do not use any further regularization beyond batch normalization since the model does not tend to overfit. \emph{Concatenation skip connections} that we employ have been shown to perform better than the popular \emph{residual skip connections} in segmentation of medical volumetric data \cite{Jensen2017}. U-net uses deconvolution layer to perform feature map upscaling but our experiments showed that bilinear interpolation performs at least as well. This is likely caused by the fact that the concatenation skip connections that were not being used when deconvolutional CNNs were first introduced provide sufficient information about fine structure to the model during upscaling. 


\subsection{Loss functions \& Optimization}
Several different multi-class loss functions used for segmentation in current literature were evaluated in this paper along with our proposed batch Dice loss. We will use the following notation to introduce different loss functions used in our experiments. Let the number of image patches $x_i$ in our training mini-batches be \(I\) and let each image patch consist of $C$ pixels. The segmentation model then maps each of $I \times C = N$ pixels in the mini-batch to probability $p_{l}$ for each of $L$ labels. The training procedure ensures that the resulting output label probability vectors $p^{c}_{l}$ correspond to one-hot encoded ground truth label vectors $r^{c}_{l}$ as best as possible on the training data. During inference, we choose the output label $l$ of each pixel $c$ as 
\begin{equation}\label{eq1}
l^{c} = \underset{l}{\arg\max}\{p_{l}^{c}\}.
\end{equation}

\textbf{Cross-Entropy}. Also known as log-loss, cross-entropy is the most widely used loss function for classification CNN. When applied to a segmentation task, cross-entropy measures the divergence of the predicted probability from the ground truth label for each pixel separately and then averages the value over all pixels in the mini-batch: 
\begin{equation}
  \label{eq2}
  \mathcal{L}_{\mathrm{CE}}=-\frac{1}{N}\sum_{c=1}^{N}\sum_{l=1}^{L}r_{l}^{c}\log{(p_{l}^{c})}
\end{equation}
This loss function tends to under-estimate the prediction probabilities for classes that are under-represented in the mini-batch which is inevitable in our training data, as can also be seen on Figure \ref{fig:ResultsOfLosses}.

\textbf{Weighted Cross-Entropy}. The tendency to under-estimate can be mitigated by assigning higher weights to loss contributions from pixels with under-represented class labels:
\begin{equation}
  \label{eq3}
  \mathcal{L}_{\mathrm{WCE}} = -\frac{1}{N}\sum_{c=1}^{N}\frac{1}{w_{c}}\sum_{l=1}^{L}r_{l}^{c}\log{(p_{l}^{c})},
\end{equation}
where $w_{c}$ is a weight assigned to pixel $c$ computed as a prior probability of ground truth label $r^c_l$ in the given mini-batch.

\textbf{Soft Dice}. Inspired by the Dice coefficient \cite{Dice1945} often used to evaluate binary segmentation accuracy, the differentiable soft Dice loss was introduced by Milletari et al.~\cite{Milletari2016} to tackle the class imbalance issue without the need for explicit weighting. One possible formulation is
\begin{equation}
  \label{eqn:LSD2}
  \mathcal{L}_{\mathrm{SD}} = 
    \frac{1}{I}\sum_{i=1}^{I}1-
    \frac{2\displaystyle\sum_{l=1}^{L}\sum_{c=1}^{C}p_{l}^{c}r_{l}^{c}}
         {\displaystyle\sum_{l=1}^{L}\sum_{c=1}^{C}p_{l}^{c}+r_{l}^{c}} .
\end{equation}
This allows easy generalization to multi-class segmentation where $L>2$ by treating each image as a 3D volume where the third dimension is the position in the one-hot encoded label vector.

\textbf{Batch Soft Dice}. We hypothesize that one of the advantages of the soft Dice loss is that it is a global operator as opposed to point-wise cross-entropy and therefore it is able to better estimate the correct overall gradient direction.  Our modification lies in extending the computation by treating the whole data mini-batch as a single 4-dimensional tensor during the loss computation. In other words, instead of computing the Dice loss over $C$ voxels $I$ times and then averaging, we compute a single Dice loss over all $N$ voxels without averaging.
\begin{equation}
  \label{eqn:LBSD}
    \mathcal{L}_{\mathrm{BSD}} =
      1-\frac{2\displaystyle\sum_{l=1}^{L}\sum_{c=1}^{N}p_{l}^{c}r_{l}^{c}}
             {\displaystyle\sum_{l=1}^{L}\sum_{c=1}^{N}p_{l}^{c}+r_{l}^{c}}
\end{equation}

Our intuition behind this choice is that during the training phase, the standard Dice loss gradient estimation on a single image/slice does not take into account the fact that the same set of filters should also be capable of segmenting structures not present in the current training slice. This is tackled by averaging the gradient over multiple slices in the batch. This can, however, cause individual gradients to more or less cancel out if their directions are very different. By contrast, computing the Dice loss gradient over the whole batch of slices as a single global operator should enforce the gradient to steadily push the filter weights towards the correct segmentation of each structure in the batch.

It should be noted that in all of the above equations we omitted regularizing term used to avoid zero division for clarity.

In the training phase, the model weights are updated through Adam optimizer with step \(10^{-4}\) computed over mini-batches of 30 image patches. As some structures are under-represented in the dataset, we use standard data augmentation techniques such as random flips, translations and elastic transformations to prevent overfitting.


\section{Experimental Results}

We optimized our model until convergence using each of the loss functions with 25 training patient scans to keep the challenge format~\cite{Raudaschl2017}. We cross-validated the models on different test and training scan subsets so that total of 10 scans were used for testing.

\begin{figure}[t]
  \centering
  \tmpframe{\includegraphics[width=1\textwidth]{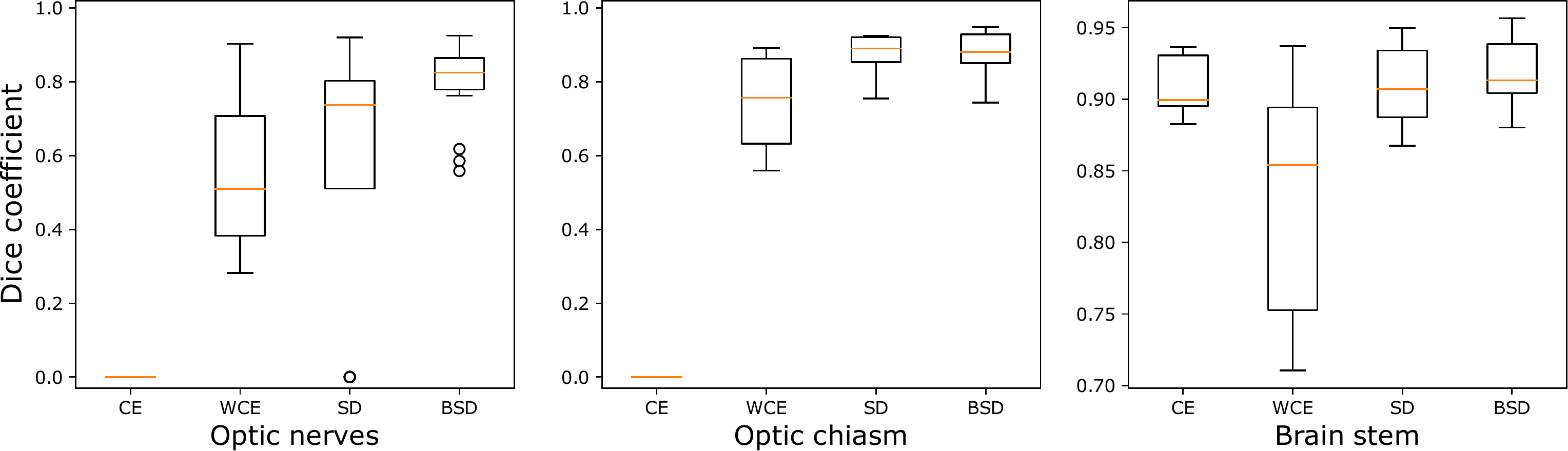}}
  \caption{Optic nerves, chiasm and brain stem segmentation results for models trained with cross-entropy (CE), weighted cross-entropy (WCE), soft Dice (SD) and batch soft Dice (BSD) loss functions. In all cases, BSD loss performs best.}
  \label{fig:ResultsOfLosses}
\end{figure}

\begin{figure}[t]
  \centering
  \tmpframe{\includegraphics[width=1\textwidth]{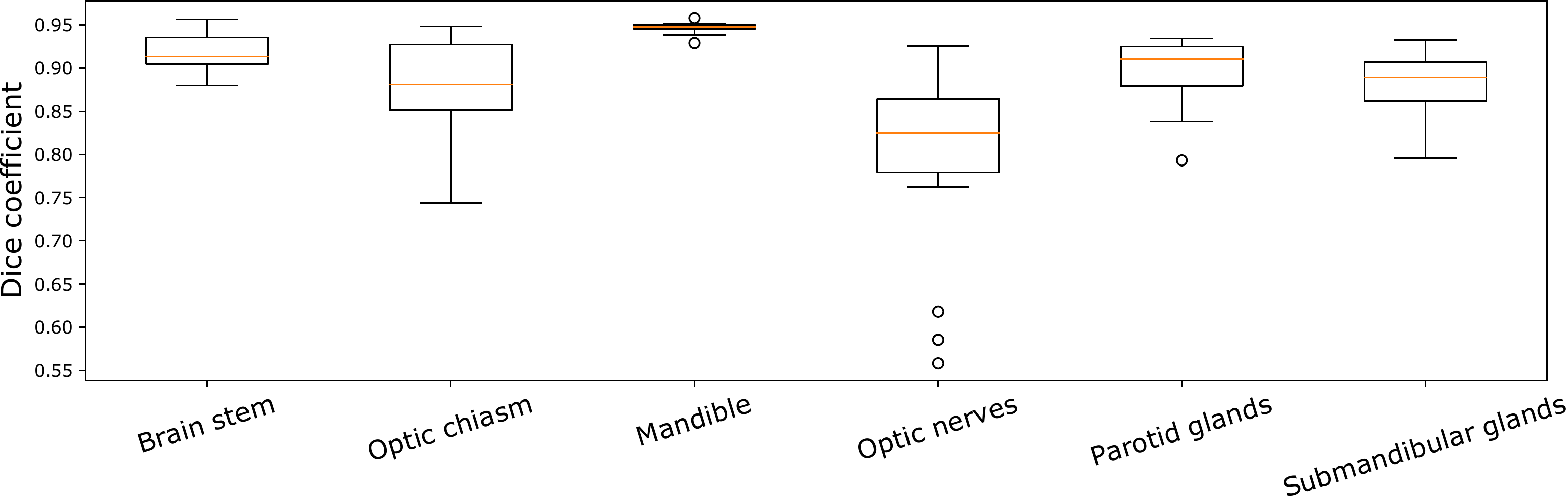}}
  \caption{Performance of segmentation model trained using $\mathcal{L}_{\mathrm{BSD}}$ loss function \eqref{eqn:LBSD} on individual OARs.}
  \label{fig:OverallResults}
\end{figure}

We first demonstrate the performance of the models on the case of optic nerves, optic chiasm, and brain stem segmentation. On other structures, the difference between performance of models trained with different loss functions is less significant. The results in terms of Dice coefficient (measure of segmentation quality on which the loss function is based) are shown in Figure~\ref{fig:ResultsOfLosses}. Superiority of soft Dice-trained models can be observed. However, standard soft Dice-trained model reaches a significantly smaller precision in case of optic nerves and sometimes misses the structure altogether as also illustrated by Figure~\ref{fig:ExamplesOfResults}. The model trained using the proposed batch soft Dice loss does not seem to suffer from this issue and we therefore conclude that it is more suitable for training models for segmentation of small anatomical structures with low contrast such as head and neck OAR.

\begin{figure}[t]
  \centering
  \tmpframe{\includegraphics[width=1\textwidth]{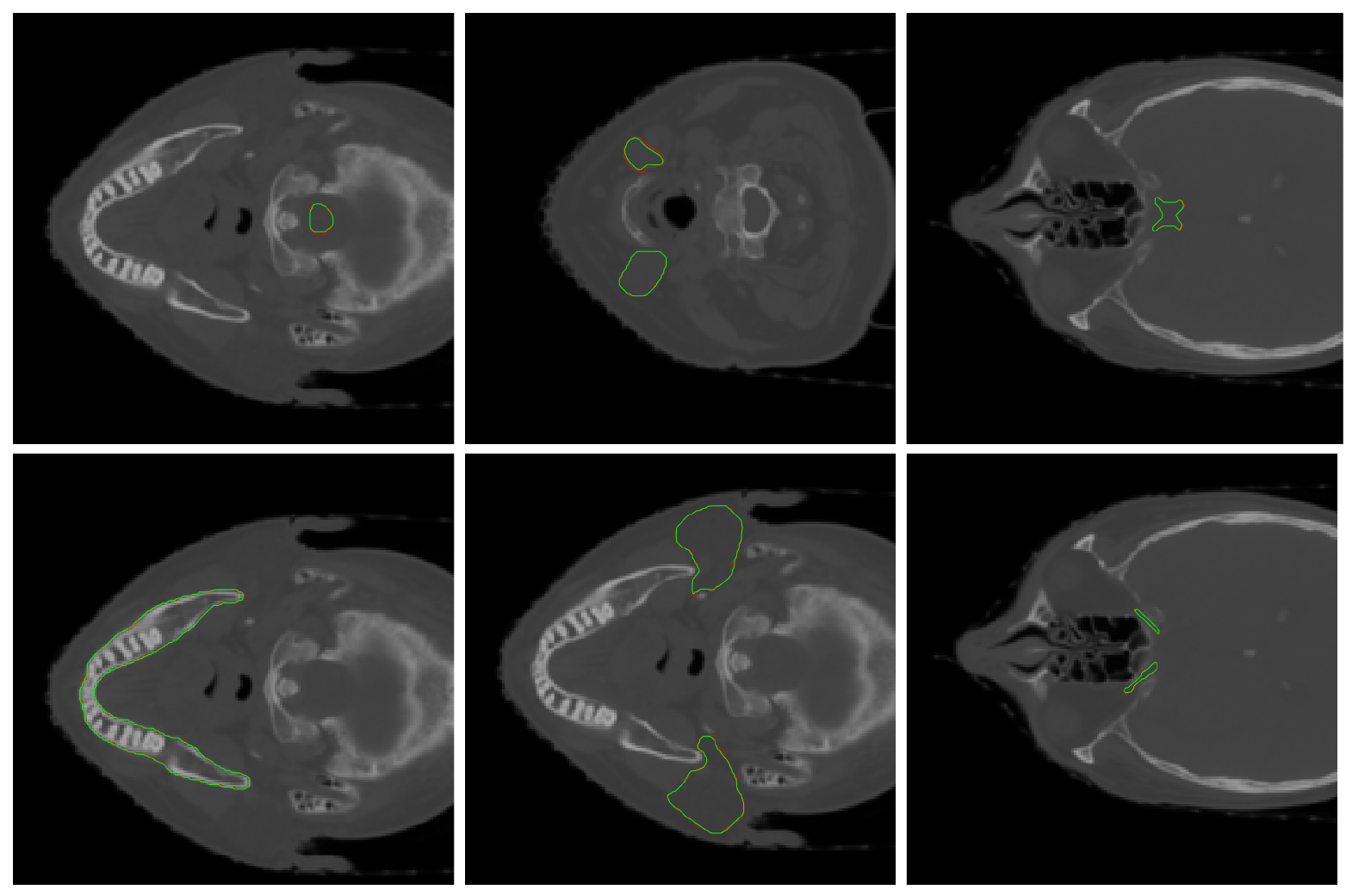}}
  \caption{Examples of successful segmentation outputs (green) with corresponding ground truth annotations (red). Upper row from left to right: brain stem, submandibular glands, optic chiasm, lower row from left to right: mandible, parotid glands, optic nerves.}
  \label{fig:segExamples}
\end{figure}

The overall performance of our model for each individual OAR is shown in Figure \ref{fig:OverallResults}. Specific examples of successful segmentation for each tissue is shown in Figure \ref{fig:segExamples}. Except for several outlier cases where optic nerve segmentation reached a lower precision with the Dice coefficient around $0.6$, we obtained acceptable results with Dice coefficient over $0.8$ for all structures. We quantitatively compare our method with other published methods in terms of Dice coefficient (Table~\ref{tab:Comparison_dice}) and in terms of average surface distance \cite{Ginneken2007} (Table~\ref{tab:Comparison_ASD}). Although the difference is most accentuated in cases of optic nerves and chiasm segmentation, our model also surpasses current state-of-the-art results on all the remaining~OARs.

\begin{table}[t]
  \centering
  \caption{Quantitative comparison of head and neck OAR segmentation methods using Dice coefficient (DSC) [\%].}
  \label{tab:Comparison_dice}
  \setlength{\tabcolsep}{3mm}
  \begin{tabular}{lcccccc}
    \midrule[0.5pt]
     \textbf{Method} & \textbf{BS} & \textbf{OC} & \textbf{MA} & \textbf{ON} & \textbf{PG} & \textbf{SG} \\ 
    \midrule[0.5pt]
    MICCAI 2015 \cite{Raudaschl2017} & 88 & 55 & 93 & 62 & 84 & 78\\
    \midrule[0.05pt]
    Fritscher et al. \cite{Fritscher2016} & - & $52$ & - & - & $81$ & $65$ \\
      & & {\scriptsize $\pm 1$} & & & {\scriptsize $\pm 4$} & {\scriptsize $\pm 8$} \\
    \midrule[0.05pt]
    Ibragimov et al. \cite{Ibragimov2017} & - & 37.4 & 89.5 & 64.2 & 77.3 & 71.4 \\
      & & {\scriptsize $\pm 13.4$} & {\scriptsize $\pm 3.6$} & {\scriptsize $\pm 7.2$} & {\scriptsize $\pm 5.8$} & {\scriptsize $\pm 11.3$} \\     
    \midrule[0.05pt]
    Wang et al. \cite{Wang2018} & 90.3 & - & 94.4 & - & 82.6 & -\\
      & {\scriptsize $\pm 3.8$} & & {\scriptsize $\pm 1.3$} & & {\scriptsize $\pm 5.7$} & \\
    \midrule[0.05pt]
    Proposed method & \textbf{91.8} & \textbf{87.9} & \textbf{94.6} & \textbf{80.0} & \textbf{89.7} & \textbf{88.1} \\
      & {\scriptsize $\pm 2.2$} & {\scriptsize $\pm 5.9$} & {\scriptsize $\pm 0.7$} & {\scriptsize $\pm 9.9$} & {\scriptsize $\pm 3.7$} & {\scriptsize $\pm 3.6$} \\
    \bottomrule
  \end{tabular}
\end{table}
\begin{table}[t]
  \centering
  \caption{Quantitative comparison of head and neck OAR segmentation methods using average surface distance (ASD) [mm].}
  \setlength{\tabcolsep}{3mm}
  \label{tab:Comparison_ASD}
    \begin{tabular}{lcccccc}
    \textbf{Method} & \textbf{BS} & \textbf{OC} & \textbf{MA} & \textbf{ON} & \textbf{PG} & \textbf{SG} \\  
    \midrule[0.5pt]
    MICCAI 2015 \cite{Raudaschl2017} & 1.1 & 1.0 & 0.5 & 0.8 & 1.6 & 1.4 \\  
    \midrule[0.05pt]
    Wang et al. \cite{Wang2018} & 0.91 & - & 0.43 & - & 1.83 & -\\
      & {\scriptsize $\pm 0.32$} & & {\scriptsize $\pm 0.12$} & & {\scriptsize $\pm 0.78$} \\
    \midrule[0.05pt]
    Proposed method & \textbf{0.63} & \textbf{0.16} & \textbf{0.29} & \textbf{0.35} & \textbf{0.82} & \textbf{1.08} \\
      & {\scriptsize $\pm 0.24$} & {\scriptsize $\pm 0.05$} & {\scriptsize $\pm 0.03$} & {\scriptsize $\pm 0.9$} & {\scriptsize $\pm 0.66$} & {\scriptsize $\pm 1.47$} \\
    \bottomrule
  \end{tabular}
\end{table}

\section{Conclusions}

We designed an encoder-decoder CNN model for head and neck OAR segmentation and proposed the Batch Dice Loss for multi-class segmentation of structures with small sizes. We compared the loss function to other standard loss functions in terms of their ability to optimize a model for OAR segmentation. The model trained using the batch Dice loss reached the best performance when compared to other loss functions and also to current state-of-the-art methods on this dataset.

In the future work we are going to evaluate the performance of batch Dice loss when applied to optimization of different models. These could include models trained on different datasets where three-dimensional models would be feasible. We are also going to assess whether it is also beneficial in the case of binary segmentation. Another potential area to explore is explicitly weighting the loss function according to current classification performance rather than prior occurrence probabilities.

\subsubsection{Acknowledgements.} This work was supported in part by the company TESCAN 3DIM (fka 3Dim Laboratory) and by the Technology Agency of the Czech Republic project TE01020415 (V3C -- Visual Computing Competence Center).

\newpage
%
%
%
%

\end{document}